\pdfoutput=1

\documentclass[11pt]{article}

\usepackage{amsmath,amsfonts,bm}

\def\eqref#1{equation~\ref{#1}}

\def\1{\bm{1}}

\def\vk{{\bm{k}}}

\def\vq{{\bm{q}}}

\DeclareMathAlphabet{\mathsfit}{\encodingdefault}{\sfdefault}{m}{sl}
\SetMathAlphabet{\mathsfit}{bold}{\encodingdefault}{\sfdefault}{bx}{n}

\usepackage[preprint]{acl}

\usepackage{times}
\usepackage{latexsym}
\usepackage{float}
\usepackage{multirow}
\usepackage[T1]{fontenc}

\usepackage{nicematrix}

\usepackage[utf8]{inputenc}

\usepackage{microtype}

\usepackage{inconsolata}

\usepackage{graphicx}

\usepackage{booktabs}
\usepackage{makecell}

\title{Quantifying Logical Consistency in Transformers via Query-Key Alignment}

\author{
  \textbf{Eduard Tulchinskii\textsuperscript{1,2}},
  \textbf{Anastasia Voznyuk\textsuperscript{3}},
  \textbf{Laida Kushnareva\textsuperscript{2}},
  \textbf{Andrei Andriiainen\textsuperscript{1,3}},
  \\
  \textbf{Irina Piontkovskaya\textsuperscript{2}},
  \textbf{Evgeny Burnaev\textsuperscript{1,5}},
  \textbf{Serguei Barannikov\textsuperscript{1,4}},
\\
\\
  \textsuperscript{1}Skolkovo Institute of Science and Technology,
  \textsuperscript{2}AI Foundation and Algorithm Lab\\
  \textsuperscript{3}Moscow Institute of Physics and Technology,
  \textsuperscript{4}CNRS, Université Paris Cité, France\\
  \textsuperscript{5}Artificial Intelligence Research Institute (AIRI)
}

\begin{document}
\maketitle

\begin{abstract}
Large language models (LLMs) have demonstrated impressive performance in various natural language processing tasks, yet their ability to perform multi-step logical reasoning remains an open challenge. Although Chain-of-Thought prompting has improved logical reasoning by enabling models to generate intermediate steps, it lacks mechanisms to assess the coherence of these logical transitions. In this paper, we propose a novel, \emph{lightweight} evaluation strategy for logical reasoning that uses query--key alignments  inside transformer attention heads. By computing a single forward pass and extracting a ``QK-score'' from carefully chosen heads, our method reveals latent representations that reliably separate valid from invalid inferences,
 offering a scalable alternative to traditional ablation-based techniques. We also provide an empirical validation on multiple logical reasoning benchmarks, demonstrating improved robustness of our evaluation method against distractors and increased reasoning depth. The experiments were conducted on a diverse set of models, ranging from 1.5B to 70B parameters.
\end{abstract}

\section{Introduction}
Large language models (LLMs) have achieved remarkable success in a range of NLP tasks, yet they still struggle with reliable multi-step logical reasoning~\cite{ChainOfThought, kojima2022large, Yang2024DoLL, Seals2023EvaluatingTD, wan-etal-2024-logicasker}. While chain-of-thought prompting has improved performance by allowing models to generate intermediate reasoning steps, it lacks a mechanism to assess the coherence of these transitions.

In this work, we propose a novel evaluation method that uses certain internal Query-Key (QK) interactions within transformer heads as a proxy for logical consistency. Specifically, given a triplet—context \(c\), statement \(s\), and candidate answer \(a_i\), $a_0 =\text{true}$ or $a_1 =\text{false}$ —we define the QK-score as:
\[
S^{(l,h)}_{QK}(c,s,a_i) = \vq_{a_i}^{(l,h)\top} \vk_{s}^{(l,h)},
\]
where \(\vq_{a_i}^{(l,h)}\) is the query vector for the answer token and \(\vk_{s}^{(l,h)}\) is the key vector corresponding to the statement, see Section \ref{sec:approach}. Our method efficiently identifies transformer heads capable of accurately evaluating the validity of logical transitions. It processes all heads in a single run, making it scalable to large models.

\begin{figure*}[t]
    \centering
      \includegraphics[width=0.99\linewidth]{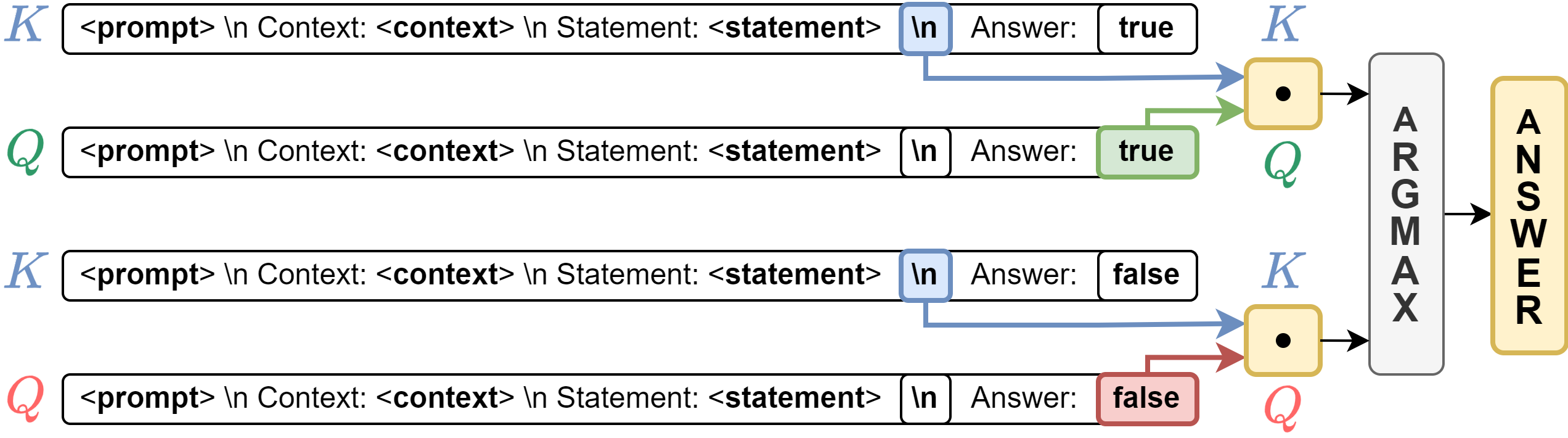}
  \caption {Our method calculates the Query-Key score between the end-of-line token immediately after the statement and the "true"/"false" tokens, for the designated head, from which we derive the answer.
}
  \label{fig:our_method}
\end{figure*}

The contributions of this paper are as follows:
\begin{itemize}
    \item We propose a novel QK-score mechanism that uses the interactions between certain query- and key-vectors to assess the correctness of logical transitions in the corresponding tasks.
    \item We conduct extensive experiments on multiple logical benchmarks to demonstrate that our method improves logical consistency, especially in multi-step reasoning scenarios or when context contain reasoning distractors.
\end{itemize}
\section{Related Work}
\label{sec:related}
A significant line of research has focused on improving multi-step logical reasoning in LLMs via chain-of-thought (CoT) prompting~\cite{ChainOfThought, kojima2022large, Yang2024DoLL, Seals2023EvaluatingTD, wan-etal-2024-logicasker}. Although CoT methods allow models to generate intermediate reasoning steps, they lack a  mechanism to assess the  coherence of these logical transitions.

Mechanistic interpretability studies have examined the roles of transformer attention heads. Recent work revealed that attention layers operate in phases—knowledge recalling, latent reasoning, and expression preparation~\cite{elhage2021mathematical, olah2020zoom, attention_heads_survey}. Subsequent studies have shown that some attention heads introduce biases by evenly splitting probabilities between answer options~\cite{lieberum2023does, yu2024correcting}, while others suppress such behaviors during the final expression phase~\cite{kim2024mechanistic}. In addition, recent investigations have attempted to analyze model behavior by disabling specific components~\cite{zhangtowards, todd2024function, NEURIPS2024_d6df31b1}, though these approaches are often computationally expensive or limited to simpler tasks~\cite{wang2023interpretability, NEURIPS2024_d6df31b1}.

For an expanded discussion on related works, see Appendix~\ref{app:more_related}.

\section{Approach}
\label{sec:approach}
Our method evaluates logical consistency by examining certain internal Query-Key (QK) interactions within transformer heads. In our setup, each input consists of a context $c$ (which provides the premises), a statement 
 $s$ (a candidate conclusion), and a candidate answer $a_i$ (with $a_0 =\text{true}$ and  $a_1 =\text{false}$), see  Figures~\ref{tab:prontoqa-eval_question_examples}, \ref{tab:pararule_plus_sample}, \ref{tab:multi-logieval_sample}  for examples.

For every attention head, indexed by \(h\) in layer \(l\), we compute a QK-score that quantifies the alignment between the representation of the statement and that of the candidate answer. Given the triplet \((c, s, a_i)\),  the \textbf{QK-score} is the dot-product
$S^{(l, h)}_{QK} (c,s,a_i) =  \vq_{a_i}^{(l, h) \top} \vk^{(l, h)}_{s}$,
where \(\vq_{a_i}^{(l,h)}\) is the query vector associated with the token representing \(a_i\) (either "true" or "false") and \(\vk_{s}^{(l,h)}\) is the key vector corresponding to the token marking the end of the statement (see Figure~\ref{fig:our_method}).  This score reflects how well a head can distinguish valid logical transitions.

By evaluating all heads in a single forward pass, our approach identifies those that reliably assess logical consistency. This efficient procedure avoids the need for extensive model modifications or head-by-head ablation studies that are common in prior work.
\begin{figure}[H]
    \centering
    \begin{tabular}{l}
        \toprule
        \makecell[l]{Use provided context and answer whether the \\ statement is true or false.} \\
        \makecell[l]{\textsc{Context}:  Each rompus is a wumpus. Every \\ rompus is not opaque. Every jompus is a \\ rompus. Every jompus is not sour. Vumpuses \\ are jompuses [...] Polly is an impus.} \\ 
        \textsc{Statement}: Polly is opaque.\\
        \textsc{Answer}: true.\\
        \bottomrule
    \end{tabular}
    \caption{PrOntoQA-OOD Example}
    \label{tab:prontoqa-eval_question_examples}
\end{figure}

\section{Experiments}
\label{sec:experimets}

\begin{figure*}
    \includegraphics[width=\linewidth]{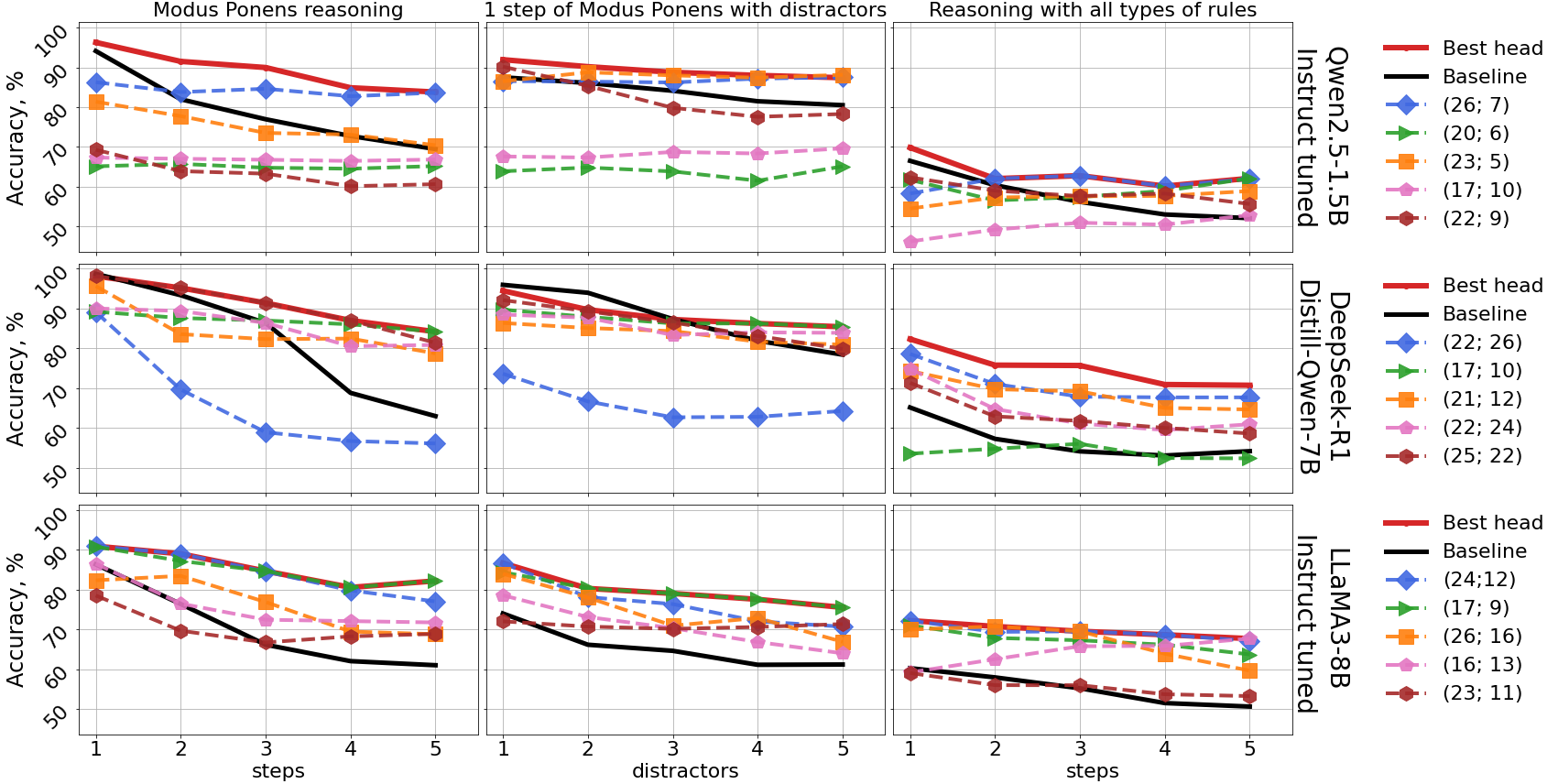}
    \caption{In-domain performamce on ProntoQA-OOD dataset. Best head was selected on calibration data for each case individually.}
    \label{fig:prontoqa_main}
\end{figure*}

\subsection{Datasets}
\label{sec:datasets}

We evaluate our method on three diverse benchmarks for logical reasoning:

\textbf{ProntoQA-OOD}~\cite{PrOntoQAOOD} is a synthetic dataset for chain-of-thought first-order logic question-answering over abstract categories (see Figure~\ref{tab:prontoqa-eval_question_examples}). We consider two deduction settings: (1) tasks requiring only repeated applications of Modus Ponens; and (2) tasks involving all six supported deduction rules. For each setup, we partition the data by reasoning depth, selecting 600 examples (300 “true” and 300 “false”) for calibration and using the remaining 1,000+ examples for evaluation. To further test robustness, we adapt the scripts to vary the number of distractors (irrelevant context statements).

\textbf{PARARULE Plus}~\cite{bao2022multi} is a dataset of true/false questions with fixed-depth reasoning. We use its test subset without modifications.

\textbf{Extended-Multi-LogiEval} builds on Multi-LogiEval~\cite{patel-etal-2024-multi}. The original dataset is constrained by only 10 samples per logical scheme and an imbalanced answer distribution. We address these limitations by generating additional samples (100 per scheme) and enforcing a balanced 50/50 split for “yes” and “no” responses (treated as \(a_0\) and \(a_1\), respectively, for QK-score computation).

All experiments are performed in a zero-shot setup. Further details are provided in Appendices~\ref{appendix:datasets_prontoqa} and \ref{appendix:datasets_mle}.

\subsection{Experimental Setup}

All experiments were performed with frozen pre-trained LLMs of size from 1.5B to 70B (for models larger than 7B, see Appendix~\ref{appnd:big_table}).

Our questions assume single-word answers; the standard approach in such setup is to select from $a_0$ and $a_1$ the option that was assigned the highest output probability by LLM, when the models are given the samples and asked to provide an answer. We refer to this method as the \textbf{Baseline}.

Our experiments were performed in two steps.
First, for each explored setup from ProntoQA-OOD, we chose the best head in terms of QK-score on calibration set. Then we report the accuracy of QK-Scoring via the chosen head and the accuracy of Baseline on the evaluation subset.

Second, we selected five heads from those that achieved the top-10 performance in various ProntoQA-OOD setups (we aimed to choose heads that cover the higher number of setups).
Then, we evaluated their performance on the {PARARULE~Plus} and {Extended-Multi-LogiEval} datasets to asses the generalization capabilities of the QK-score informed head selection.

\begin{table*}[h!]
    \centering
    \begin{NiceTabular}{rccccc|cccc}
        & & \multicolumn{4}{c}{\textbf{PARARULE Plus}} & \multicolumn{4}{c}{\textbf{Extended-Multi-LogiEval}} \\
        &   & \multicolumn{4}{c}{\textbf{Reasoning depth}} & \multicolumn{4}{c}{\textbf{Reasoning depth}} \\
        
        & Head & \textbf{2} & \textbf{3} & \textbf{4} & \textbf{5} & \textbf{1} & \textbf{2} & \textbf{3} & \textbf{4}\\
        \cmidrule(lr){2-10} 
         Qwen2.5 & (26, 7) &  0.6043 & \underline{0.6772} & 0.6483 & \underline{0.7095} & 0.4730 & 0.4663 & \underline{0.5163} & 0.4413 \\
          1.5B, instruct & (20, 6) & {0.5310} & {0.5844} & {0.5436} & 0.6097 &  \textbf{0.6069} & \textbf{0.6063} & \textbf{0.5804} & \textbf{0.5516} \\
          & (23, 5) & 0.5999 & \underline{0.6488} & 0.6468 & \underline{0.6694} & {0.5026} & \underline{0.5587} & 0.4859 & 0.3840\\
          & (17, 10) & 0.4552 & 0.5569 & 0.4800 & 0.4549 & {0.5003} & \underline{0.5165} & 0.5413 & 0.4943 \\
          & (22, 9) & \textbf{0.7095} & \textbf{0.7529} & \textbf{0.7719} & \textbf{0.8003} & 0.2445 & 0.4531 & 0.4141 & 0.4527 \\
           \multicolumn{1}{r}{Baseline} & -  & 0.6240 & 0.6263 & 0.6521 & 0.6423 & 0.4996 & 0.4834 & 0.5011 & 0.5057  \\
          \midrule %
          DeepSeek-R1 & (22, 26) & \textbf{0.6736} & \textbf{0.7418} & \underline{0.5495} & \underline{0.6145} & 0.5236   & \underline{0.5112} & \underline{0.5402} & \underline{0.5115} \\
          \small{Distill-Qwen-7B} & (22, 16) & 0.3206 & 0.3422  & 0.2763 & 0.2388 & 0.4689     & \underline{0.5271}     & \underline{0.5185}     & \underline{0.5143} \\
          & (21, 12) &  0.4982 &  0.5203 & 0.4452 & 0.4740 & \textbf{0.7682}     & \textbf{0.5733}     & \textbf{0.5989} & \textbf{0.6218} \\
          & (22, 24) & \underline{0.6523} & \underline{0.6454} & \textbf{0.6149} & \textbf{0.6557} & \underline{0.6572} & \underline{0.5495}  & \underline{0.5771} & \underline{0.5888} \\
          & (25, 22) & \underline{0.6227} & \underline{0.6233} & \underline{0.6145} & 0.5357  &  0.5093 & 0.4768  & 0.4902 & \underline{0.5186} \\
          \multicolumn{1}{r}{Baseline} &  - & {0.4969} & {0.5212} & 0.4523 & 0.4717 & 0.5153     & 0.4835     & 0.5010 & 0.5057\\
    \end{NiceTabular}
    \caption{Performance of QK-score on heads selected on ProntoQA-OOD in cross-domain evaluation. PARARULE Plus and Extended-Multi-LogiEval datasets are used. Best results are highlighted in \textbf{bold}. Those results that are better than baseline are \underline{underlined}}
    \label{tab:pronto_to_pararulel}
\end{table*}

\section{Results}
\label{sec:results}

\subsection{In-Domain Evaluation}
For each deduction setup in ProntoQA-OOD, we select the best-performing head on the calibration set (referred to as \textsc{Best Head}) and evaluate its performance on the held-out evaluation set. Figure~\ref{fig:prontoqa_main} shows that the \textsc{Best Head} consistently outperforms the baseline across varying reasoning depths and under different numbers of distractors. In addition, we report the performance of five individual heads drawn from the top-five performers in each setup. These results indicate that the selected heads maintain stable performance regardless of the number of reasoning steps, thereby confirming that our QK-score reliably identifies transformer components that contribute to logical consistency. Results for three LLMs are presented here; further details are provided in Appendix~\ref{appnd:big_table}.)

\subsection{Transfer Learning Evaluation}
We further assessed the cross-dataset generalization of our approach by evaluating the QK-scoring accuracy of the heads selected in the above experiments on two additional datasets: PARARULE Plus and Extended-Multi-LogiEval. As reported in Table~\ref{tab:pronto_to_pararulel}, three out of five selected heads achieve an accuracy exceeding the baseline by more than 10\% on PARARULE Plus. In contrast, only two heads outperform the baseline on Extended-Multi-LogiEval, which may be attributed to differences in question format (binary "yes/no" vs. true/false evaluations). 

Overall, these results demonstrate that our QK-score method  identifies transformer heads capable of reliably assessing logical transitions, with robust performance observed both within the ProntoQA-OOD domain and in cross-dataset setups.

\section{Analysis}

Our findings suggest that QK-scores offer a distinctive lens on how LLMs process logical structure. Unlike raw attention weights, QK-scores are independent of positional encodings, thereby focusing purely on semantic alignment between the statement and candidate answers. Moreover, heads selected via QK-scores often outperform the model’s final probabilities, confirming that essential reasoning signals can be sometimes obscured by later-stage processing~\cite{kim2024mechanistic, lieberum2023does}.

We also observe that high QK-scoring heads consistently identify valid inferences even with distractors in the input. This stability indicates that such heads act as “verification anchors,” largely unaffected by irrelevant context. Consequently, QK-scores may bolster both interpretability and robustness: by highlighting the heads that preserve logical consistency, they help clarify how multi-step reasoning unfolds within large language models.

\section{Conclusion}We introduced a QK-score framework for uncovering attention heads that consistently capture logical validity. Our experiments show that in multi-step inferences with distractors, certain heads outperform the model’s own final-layer predictions. Crucially, this single-pass procedure sidesteps the computational overhead of ablation-based methods and generalizes relatively well across datasets. By identifying attention heads that act as “reasoning checkpoints,” our approach offers a tractable window into how these models process complex logical relationships. Future work may refine QK-scoring for specialized tasks, explore synergy with chain-of-thought prompting, and extend this analysis to other interpretability settings. Ultimately, bridging internal alignment signals with logical consistency represents a key step toward transparent and reliable language models.

\section{Limitations}

Application of our method requires a sufficiently large calibration dataset (at least $\approx 400$ reasoning questions). This dataset must cover logical rules with more-or-less balanced number of examples and must be carefully debiased to prevent selection of attention heads that guess the answer from the form of the question.

We do not state that heads identified by our method are the only ones responsible for the logical inference within the model.

During the research for this short paper, we performed head selection only on two sets of deduction rules ("Modus Ponens" vs. "Modus Ponens, conjunction/disjunction introduction/elimination, and proof by contradiction"). It is quite possible, that the `scope' of certain attention heads is limited to different sets of logical principles, and a more extensve experiments may be needed to explore those cases.

Finally,  we considered only Instruct- and Chat-tuned models. Question of if and how capabilities for logic in attention heads change during the transition from base to fine-tuned version we leave for the future work.

\bibliography{main}

\appendix

\section{Related Work}
\label{app:more_related}
Here we discuss in more details the works related to logical reasoning and internal model interpretability.

\textbf{Chain-of-Thought (CoT) and Logical Reasoning.}
CoT prompting has been widely adopted to improve reasoning in LLMs~\cite{ChainOfThought, kojima2022large, Yang2024DoLL, Seals2023EvaluatingTD, wan-etal-2024-logicasker}. These methods prompt models to produce intermediate steps in the reasoning process. However, despite the success in various tasks, they do not offer  a  measure to quantify the coherence of the reasoning transitions, leaving a gap that our QK-score method aims to fill.

\textbf{Mechanistic Interpretability.}
A complementary line of research has focused on understanding transformer models through mechanistic interpretability. Work by \citet{elhage2021mathematical} and \citet{olah2020zoom} has shown that transformer attention heads can be categorized into different functional phases, including knowledge recalling, latent reasoning, and expression preparation~\cite{attention_heads_survey}. More recent studies have explored specific reasoning tasks: for example, \citet{kim2024mechanistic} examined syllogistic reasoning, showing that models learn content-independent reasoning mechanisms transferable across different logical schemes. Other investigations~\cite{lieberum2023does, yu2024correcting} have highlighted that certain heads can adversely affect the final decision by introducing latent biases. While ablation-based techniques~\cite{zhangtowards, todd2024function, NEURIPS2024_d6df31b1} have been used to study these phenomena, they are often resource-intensive or limited to smaller models~\cite{wang2023interpretability}. 

\textbf{Benchmark Datasets.}
Several benchmarks have been designed to evaluate logical reasoning in LLMs. For example, LogicBench~\cite{parmar-etal-2024-logicbench} and Multi-LogiEval~\cite{patel-etal-2024-multi} test models on tasks such as truth table reasoning, logical entailment, and satisfiability. Datasets specifically tailored for chain-of-thought evaluation, such as PrOntoQA~\cite{prontoqa}, FOLIO~\cite{han2022folio}, and FLD~\cite{morishita_2024_NeurIPS_FLD_diverse}, demonstrate that unstructured intermediate reasoning steps can enhance performance.

\textbf{Alternative Evaluation Strategies.}
Other lines of work have explored neuro-symbolic and data-driven methods to assess reasoning quality. Some approaches reformulate reasoning tasks into structured formats, while others propose direct evaluation metrics based on model outputs. 

\textbf{Summary.}
While previous work has substantially advanced our understanding of how LLMs reason and how their internal representations operate, there remains a need for efficient, scalable methods to assess logical consistency. Our method, which relies on the natural alignment between specific query and the last statement token key vectors, complements existing techniques and offers an efficient alternative to ablation-based analysis.

\section{ProntoQA-OOD: additional details}
\label{appendix:datasets_prontoqa}

Questions in PrOntoQA-OOD are organized in a following way: given a set of axioms (context) it is required to prove a theorem (statement). For our study we reformulate them into answering if the statement is true or false given the context.

We used scripts publsihed by the authors of the dataset to generate the data. We used following command line flags:

\begin{itemize}
    \item For Modus Ponense only: \\
    {\texttt{-{}-ordering random -{}-distractors none -{}-deduction-rule ModusPonens -{}-proofs-only -{}-ontology fictional -{}-min-hops 1 -{}-max-hops 5}} 
    \item For composition of deduction rules: \\ \texttt{-{}-ordering random  -{}-deduction-rule Composed -{}-proofs-only -{}-distractors none -{}-ontology fictional  -{}-min-hops 1 --max-hops 5}
\end{itemize}

We also modified the original scripts to make possible variating the number of distractors in prompts, and generated data for $1-$hop questions on Modus Ponense with $1 , 5$ distractors respectively. 

In our experiments we used training data from \textit{in context examples}. For each setup (deduction rule$ + $number of hops $+$ number of distractors) scripts yielded $4,000$ samples. To exclude possible biases, in each case we select equal number of questions where the statement is given in positive (\texttt{X is Y.}) and negative (\texttt{X is not Y.}) forms and for each sample we also generate its counterpart where its statement and ground-truth answer are negated (thus balancing the classes). Depending on the setup, this resulted in $2,600 - 7,000$ questions, out of which 600 ($300/300$ with answers "true" and "false" respectively) were taken for calibration and the rest were used as evaluation set (see Table~\ref{tab:samples_pronto}). We ensured that no pair axioms$+$theorem is included in both subsets.

\begin{table}[]
    \addtolength{\tabcolsep}{-0.2em}
    \centering
    \begin{NiceTabular}{l|ccccc}
         \textbf{Deduction} & \multicolumn{5}{c}{\textbf{Number of hops}} \\
         \textbf{Rule} & \textbf{1} & \textbf{2} & \textbf{3} & \textbf{4} & \textbf{5}  \\
         \midrule
         M. P. & 6,144 & 6,204 & 6,364 & 6,288 & 6,176\\
        \makecell[l]{M. P. \\+ distr.} & {6,268} & - & - & - & - \\
         Comp. & 2,752 & 2,764 & 2,884 & 2,956 & 2,892\\
    \end{NiceTabular}
    \addtolength{\tabcolsep}{0.2em}
    \caption{Size of ProntoQA-OOD evaluation set that was used in different setups}
    \label{tab:samples_pronto}
\end{table}

\section{Pararule-plus: additional details}

Here we demonstrate the example from this dataset:

\begin{figure}[H]
    \centering
    \begin{tabular}{l}
        \toprule

        \makecell[l]{\textsc{Context}: Harry is strong. Harry is big. Harry \\is high.  Anne is thin. Anne is little. Gary is \\smart. Gary is quiet. Gary is kind. Fiona is poor.\\Fiona is rough. Fiona is sad. Strong people are \\smart. If someone is thin and little then they are\\short. If someone is poor and rough then they\\are bad. If someone is smart and quiet then they\\are nice. All short people are small. All smart\\people are quiet. All nice people are wealthy.\\ All bad people are dull.} \\ 
        \textsc{Statement}: Harry is quiet. \\
        \textsc{Answer}: true\\
        \bottomrule
    \end{tabular}
    \caption{PARARULE-PLUS prompt example of reasoning, depth 2}
    \label{tab:pararule_plus_sample}
\end{figure}

On {PARARULE Plus}, three out of five heads, selected on the ProntoQA-OOD dataset, reach accuracy higher than the baseline, in most cases by more than $10\%$. Interestingly, head \texttt{(22, 16)} from DeepSeek-R1 consistently yields an accuracy below 0.35 on PARARULE Plus, suggesting that its QK-score distinguishes correct from incorrect logical implications but in a reversed manner. 
Similar effect also occurs in some setups on other heads.

\section{Extended Multi-Logi-Eval: additional details}
\label{appendix:datasets_mle}

Here we demonstrate the example from this dataset and how it was prompted:
\begin{figure}[H]
    \centering
    \begin{tabular}{l}
        \toprule
        \makecell[l]{Use provided Context to answer the Question. \\ Print 'yes' or 'no' only.} \\ 
        \makecell[l]{\textsc{Context}: If a person uses a fishing rod, \\ they catch fish. Michael uses a fishing rod.} \\ 
        \textsc{Question}: Does Michael catch fish?\\
        \textsc{Answer}: yes.\\
        \bottomrule
    \end{tabular}
    \caption{Multi-LogiEval Modus Ponens Example of reasoning depth 1}
    \label{tab:multi-logieval_sample}
\end{figure}

While the original dataset Multi-Logi-Eval consisted of three types of logic, namely First-Order Logic, Nonmonotonic Logic and Propositional Logic, we extended only a part with first-order logic. We used GPT-4o to generate more samples for each scheme.
Table~\ref{tab:statistics} compares the statistics of generated dataset with the statistics of the original (Multi-LogiEval dataset):
\begin{table}[H]
\begin{tabular}{lcccc}
& \multicolumn{4}{c}{\textbf{Depth}} \\ 
\cmidrule(lr){2-5} 
\textbf{Dataset}   & \multicolumn{1}{c}{1}    & \multicolumn{1}{c}{2}   & \multicolumn{1}{c}{3}   & 4   \\ \hline
\makecell[l]{Multi-LogiEval \\ (FOL)} & \multicolumn{1}{l}{130}  & \multicolumn{1}{l}{105} & \multicolumn{1}{l}{135} & 120 \\ \hline
\makecell[l]{Extended \\ Multi-LogiEval} & \multicolumn{1}{l}{1300} & \multicolumn{1}{l}{700} & \multicolumn{1}{l}{900} & 700 \\ \bottomrule
\end{tabular}
\caption{Statistics of generated Extended-MultiLogiEval dataset.}
\label{tab:statistics}
\end{table}

We refer to original paper~\cite{patel-etal-2024-multi} for the detailed description of logical schemes, and we did not add any new schemes or changed them in any way on all levels of reasoning. Every scheme of reasoning depth $k$ consists of $k$ atomic rules from reasoning depth 1, see them in Table~\ref{tab:classical_inference_rules}.
\begin{table*}[htbp]
  \centering
    \begin{tabular}{c|c} \toprule
    \textbf{Rule} & \textbf{First-order Logic} \\
    \midrule
    MP & $(\forall x(p(x) \to q(x)) \land p(a)) \vdash q(a) $ \\  \midrule
    
    MT & $(\forall x(p(x) \to q(x)) \land \neg q(a)) \vdash \neg p(a) $ \\ \midrule
    
    HS & $(\forall x((p(x) \to q(x)) \land (q(x) \to r(x))) \vdash (p(a) \to r(a)) $ \\ \midrule
    
    DS & $(\forall x(p(x) \lor q(x)) \land \neg p(a)) \vdash q(a) $  \\ \midrule
    
    CD & $(\forall x((p(x) \to q(x)) \land  (r(x) \to s(x))) \land (p(a) \lor r(a)))\ \vdash (q(a) \lor s(a)) $ \\ \midrule
    
    DD & $(\forall x((p(x) \to q(x)) \land  (r(x) \to s(x))) \land (\neg q(a) \lor \neg s(a)))\ \vdash (\neg p(a) \lor \neg r(a)) $ \\ \midrule
    
    BD & $(\forall x((p(x) \to q(x)) \land  (r(x) \to s(x))) \land (p(a) \lor \neg s(a)))\ \vdash (q(a) \lor \neg r(a)) $ \\ 
    \midrule
    
    CT & $\forall x(p(x) \lor q(x)) \dashv\vdash \forall x(q(x) \lor p(x))$ \\
    \midrule
    DMT  & $\neg\forall x (p(x) \land q(x)) \dashv\vdash \exists x(\neg p(x) \lor \neg q(x))$ \\
    \midrule
    CO & $\forall x ((p(x)\to q(x)) \land (p(x)\to r(x)))\vdash \forall x (p(x) \to (q(x) \land r(x)))$ \\
    \midrule
    IM & $\forall x(p(x)\to (q(x)\to r(x)))\dashv\vdash \forall x((p(x)\land q(x)) \to r(x))$ \\
    \midrule
    EG  &  $p(a) \vdash \exists x(p(x))$ \\
    \midrule
    UI &  $ \forall x(p(x)) \vdash p(a)$ \\
    
    \bottomrule
    \end{tabular}%
    \caption{Inference rules that establish the relationship between premises and their corresponding conclusions in Extended Multi-LogiEval. The schemes are MP: Modus Ponens, MT: Modus Tollens, HS: Hypothetical Syllogism, DS: Disjunctive Syllogism, CD: Constructive Dilemma, DD: Destructive Dilemma, BD: Bidirectional Dilemma, CT: Commutation, DMT: De Morgan's Theorem, CO: Composition, IM: Importation, EG: Existential Generalization, UI: Universal Instantiation}
    \label{tab:classical_inference_rules}%
\end{table*}

\section{Extended results of in-domain evaluation on ProntoQA-OOD}
\label{appnd:big_table}
Tables \ref{tab:prontoqa_depth} and \ref{tab:prontoqa_distractors} provide the in-domain numerical evaluation results on ProntoQA-OOD for \textsc{Best head} and \textsc{Baseline} methods calculated for various LLMs.

\begin{table*}[ht]
    \addtolength{\tabcolsep}{-0.2em}
    \centering
    \begin{tabular}{llccccc| ccccc}
        \toprule
        & & \multicolumn{5}{c}{\textbf{Modus ponens only}} & \multicolumn{5}{c}{\textbf{Composition of rules}} \\
        \cmidrule(lr){3-12}
        \multicolumn{2}{r}{\textbf{Reasoning depth: }} & \textbf{1} & \textbf{2} & \textbf{3} & \textbf{4} & \textbf{5} & \textbf{1} & \textbf{2} & \textbf{3} & \textbf{4} & \textbf{5}\\
        \midrule
        \textbf{LLaMA2} & QK & \textbf{0.8889} & \textbf{0.8427} & \textbf{0.8474} & \textbf{
        0.8524} & \textbf{0.8444} & \textbf{0.7474} & \textbf{0.7847} & \textbf{0.7237} &	\textbf{0.7013} & \textbf{0.7004} \\
        \textbf{7B Chat} & Baseline & 0.6772 & 0.6349 & 0.6506 & 0.6197 & 0.6072 & 0.5678 & 0.5414 & 0.5427 & 0.5232 & 0.5346 \\
        \midrule
        \textbf{LLaMA2} &       QK   & \textbf{0.9778} & \textbf{0.9637} & \textbf{0.9283} & \textbf{0.9187} & \textbf{0.9107} & \textbf{0.7584} & \textbf{0.7134} & \textbf{0.7044} & \textbf{0.6777} & \textbf{0.6964} \\
        \textbf{13B Chat} & Baseline & 0.8944          & 0.7784          & 0.7527 & 0.7385 & 0.7414 & 0.6556 & 0.4944 & 0.4932 & 0.4751 & 0.4786 \\
        \midrule
        \textbf{LLaMA3} & QK & \textbf{0.9090} & \textbf{0.8899} & \textbf{0.8465} & \textbf{0.8054} & \textbf{0.8217} & \textbf{0.7219} & \textbf{0.7073} & \textbf{0.6951} & \textbf{0.6863} & \textbf{0.6767} \\
        \textbf{{8B Instruct}} & Baseline & 0.8632 & 0.7636 & 0.6613 & 0.6199 & 0.6097 & 0.6023 & 0.5791 & 0.5517 & 0.5145 & 0.5057 \\
       \midrule
        \textbf{LLaMA3.1} &       QK    & \textbf{0.9881} & 0.9577 & 0.9192 & 0.8799 & 0.8676 & \textbf{0.8084} & \textbf{0.7184} & \textbf{0.7188} & \textbf{0.7239} & \textbf{0.7052} \\
        \textbf{8B Instruct} & Baseline & 0.9843 & \textbf{0.9785} & \textbf{0.9473} & \textbf{0.9035} & \textbf{0.8775} & 0.6848 & 0.6206 & 0.5847 & 0.5888 & 0.5731 \\
        \midrule
        \textbf{LLaMA3} &       QK       & \textbf{1.0000} & \textbf{0.9999} & 0.9963 & \textbf{0.9981} & \textbf{0.9960} & 0.\textbf{9890} & \textbf{0.9214} & \textbf{0.8799} & \textbf{0.8617} & \textbf{0.8353} \\
        \textbf{70B Instruct} & Baseline & 0.9993 & \textbf{0.9999} & \textbf{0.9981} & 0.9868 & 0.9739 & 0.9805 & 0.8767 & 0.8324 & 0.8171 & 0.7948 \\
        \midrule
        \textbf{LLaMA3.1} &       QK     & 0.9987 & 0.9993 & \textbf{0.9993} & \textbf{0.9968} & \textbf{0.9942} & 0.9890 & \textbf{0.9144} & \textbf{0.9073} & \textbf{0.8762} & \textbf{0.8631} \\
        \textbf{70B Instruct} & Baseline & \textbf{1.0000} & \textbf{0.9999} & 0.9967 & 0.9904 & 0.9762 & \textbf{0.9922} & 0.8912 & 0.8685 & 0.8348 & 0.8246 \\
        \midrule
        \textbf{Qwen-2.5} & QK & \textbf{0.9627} & \textbf{0.9149} & \textbf{0.8995} & \textbf{0.8487} & \textbf{0.8379} & \textbf{0.6981} & \textbf{0.6201} & \textbf{0.6275} & \textbf{0.6135} & \textbf{0.6209} \\
        \textbf{{1.5B Instruct}} & Baseline & 0.9410 & 0.8197 & 0.7694 & 0.7273 & 0.6946 & 0.6648 & 0.6036 & 0.5617 & 0.5299 & 0.5207  \\
        \midrule
        \textbf{Qwen-2.5} &       QK     & \textbf{0.9988} & \textbf{0.9966} & \textbf{0.9869} & \textbf{0.9791} & \textbf{0.9683} & \textbf{0.9724} & \textbf{0.8813} & \textbf{0.8313} & \textbf{0.8332} & \textbf{0.8397} \\
        \textbf{14B Instruct} & Baseline & 0.9660 & 0.9656 & 0.9388 & 0.9146 & 0.9060 & 0.9050 & 0.8374 & 0.7720 & 0.7323 & 0.7451 \\
        \midrule
        \textbf{Qwen-2.5} &       QK     & \textbf{1.0000} & \textbf{0.9996} & \textbf{0.9947} & \textbf{0.9929} & \textbf{0.9916} & \textbf{0.9543} & \textbf{0.9257} & \textbf{0.8598} & \textbf{0.8340} & \textbf{0.8104} \\
        \textbf{32B Instruct} & Baseline & 0.9988 & 0.9975 & 0.9783 & 0.9203 & 0.8502 & 0.9274 & 0.8623 & 0.8264 & 0.7982 & 0.7710 \\
        \midrule
        \textbf{DeepSeek} & QK & \textbf{0.9149} & \textbf{0.8039} & \textbf{0.8995} & \textbf{0.8487} & \textbf{0.8379} & \textbf{0.6718} & \textbf{0.6393} & \textbf{0.5976} & \textbf{0.5825} & \textbf{0.5894} \\
        \scriptsize{\textbf{R1-Distill-Qwen-1.5B}} & Baseline & 0.5361 & 0.5103 & 0.5053 & 0.5011 & 0.5020 & 0.4873 & 0.4980 & 0.4959 & 0.4971 & 0.4985 \\
        \midrule
        \textbf{DeepSeek} & QK & {0.9816} & \textbf{0.9519} & \textbf{0.9141} & \textbf{0.8701} & \textbf{0.8418} & \textbf{0.8233} & \textbf{0.7577} & \textbf{0.7568} & \textbf{0.7090} & \textbf{0.7071} \\
        \scriptsize{\textbf{R1-Distill-Qwen-7B}} & Baseline & \textbf{0.9873} & 0.9336 & 0.8638 & 0.6879 & 0.6295 & 0.6512 & 0.5723 & 0.5408 & 0.5304 & 0.5413 \\
        \bottomrule
    \end{tabular}
    \addtolength{\tabcolsep}{0.2em}
    \caption{Comparison of models with different reasoning depths on ProntoQA-OOD. Best results are highlighted in bold.}
    \label{tab:prontoqa_depth}
\end{table*}

\begin{table*}[ht]
    \centering
    \begin{tabular}{llccccc}
        \toprule
        & & \multicolumn{4}{c}{\textbf{Distractors added}} \\
        \cmidrule(lr){3-7}
        & & \textbf{1} & \textbf{2} & \textbf{3} & \textbf{4} & \textbf{5} \\
        \midrule
        \textbf{LLaMA2} & QK & \textbf{0.8725} & \textbf{0.8654} & \textbf{0.8595} & \textbf{0.8487} & \textbf{0.8498} \\
        \textbf{7B Chat} & Baseline & 0.6577 & 0.6583 & 0.6422 & 0.6280 & 0.6248 \\
        \midrule
        \textbf{LLaMA2} &       QK       & \textbf{0.9627} & \textbf{0.8972} & \textbf{0.8981} & \textbf{0.8920} & \textbf{0.8552} \\
        \textbf{13B Chat} & Baseline & 0.7959 & 0.7348 & 0.7160 & 0.7005 & 0.6677 \\
        \midrule
        \textbf{LLaMA3} & QK & \textbf{0.8665} & \textbf{0.8028} & \textbf{0.7902} & \textbf{0.7753} & \textbf{0.7551}  \\
        \textbf{8B Instruct} & Baseline & 0.8632 & 0.7636 & 0.6613 & 0.6199 & 0.6097  \\
       \midrule
        \textbf{LLaMA3.1} &       QK    & 0.9527 & 0.8577 & \textbf{0.8170} & \textbf{0.7895} & \textbf{0.7699} \\
        \textbf{8B Instruct} & Baseline & \textbf{0.9629} & \textbf{0.8691} & 0.7830 & 0.7212 & 0.6672 \\
        \midrule
        \textbf{LLaMA3} &       QK       & \textbf{0.9967} & \textbf{0.9947} & \textbf{0.9897} & \textbf{0.9849} & \textbf{0.9845} \\
        \textbf{70B Instruct} & Baseline & 0.9961 & 0.9882 & 0.9463 & 0.9106 & 0.8926 \\
        \midrule
        \textbf{LLaMA3.1} &       QK     & \textbf{0.9971} & \textbf{0.9954} & \textbf{0.9853} & \textbf{0.9672} & \textbf{0.9613} \\
        \textbf{70B Instruct} & Baseline & 0.9933 & 0.9918 & 0.9582 & 0.9176 & 0.8916 \\
       \midrule
        \textbf{Qwen-2.5} & QK & \textbf{0.9627} & \textbf{0.9149} & \textbf{0.8995} & \textbf{0.8487} & \textbf{0.8379} \\
        \textbf{1.5B Instruct} & Baseline & 0.9410 & 0.8197 & 0.7694 & 0.7273 & 0.6946 \\
        \midrule
        \textbf{Qwen-2.5} &       QK     & \textbf{0.9955} & \textbf{0.9823} & \textbf{0.9556} & \textbf{0.9362} & \textbf{0.9320} \\
        \textbf{14B Instruct} & Baseline & 0.9687 & 0.9528 & 0.8896 & 0.8336 & 0.8130 \\
        \midrule
        \textbf{Qwen-2.5} &           QK & \textbf{0.999}7 & \textbf{0.9973} & \textbf{0.9794} & \textbf{0.9546} & \textbf{0.9429} \\
        \textbf{32B Instruct} & Baseline & 0.9990 & 0.9931 & 0.9253 & 0.8042 & 0.7522 \\
        \midrule
        \textbf{Deepseek} & QK & \textbf{0.8290} & \textbf{0.7650} & \textbf{0.7485} & \textbf{0.7574} & \textbf{0.7501} \\
        \scriptsize{\textbf{R1-Distill-Qwen-1.5B}} & Baseline & 0.5361 & 0.5096 & 0.5098 & 0.5016 & 0.5012 \\
        \midrule
        \textbf{Deepseek} & QK & {0.9448} & {0.8972} & {0.8725} & \textbf{0.8625} & \textbf{0.8540} \\
        \scriptsize{\textbf{R1-Distill-Qwen-7B}} & Baseline & \textbf{0.9597} & \textbf{0.9397} & \textbf{0.8736} & 0.8191 & 0.7840 \\
        \bottomrule
    \end{tabular}
    \caption{Effect of distractors added to the prompt on ProntoQA-OOD. Only Modus Ponens inference.}
    \label{tab:prontoqa_distractors}
\end{table*}

\end{document}